\pdfoutput=1

\documentclass[11pt]{article}

\usepackage[final]{acl}

\usepackage{times}
\usepackage{latexsym}

\usepackage{booktabs}       
\usepackage{microtype}      
\usepackage{booktabs}       
\usepackage{url}            
\usepackage{amsfonts}       
\usepackage{nicefrac}       
\usepackage{amsmath}
\usepackage{graphicx}
\usepackage{bm, upgreek}
\usepackage{amssymb}
\usepackage{array}
\usepackage{multirow}
\usepackage{mathtools}
\usepackage{arydshln}
\usepackage{tabularx}
\usepackage{color}
\usepackage{soul}
\usepackage{caption}
\usepackage{xcolor}
\usepackage{comment}
\usepackage{adjustbox}
\usepackage{hyperref}
\usepackage{cleveref}
\usepackage[ruled,vlined]{algorithm2e}
\usepackage{makecell}
\usepackage{diagbox}
\usepackage{subfigure}
\usepackage{subcaption}
\usepackage{xcolor, soul}
\usepackage{wrapfig}

\usepackage{pifont}
\definecolor{bluecolor}{HTML}{0000FF}
\definecolor{greencolor}{HTML}{8CD0A4}
\definecolor{yellowcolor}{HTML}{F9D17C}
\definecolor{redcolor}{HTML}{FF0000}

\definecolor{black}{rgb}{0,0,0}

\usepackage[T1]{fontenc}

\usepackage[utf8]{inputenc}

\usepackage{microtype}

\usepackage{inconsolata}

\usepackage{graphicx}

\makeatletter
\def\@fnsymbol#1{\ensuremath{\ifcase#1\or \dagger \or  \ddagger\or
   \mathsection\or  \text{*}\or \mathparagraph \or  \| \or **\or \dagger\dagger
   \or \ddagger\ddagger \else\@ctrerr\fi}}
\makeatother
 
\renewcommand{\thefootnote}{\fnsymbol{footnote}}

\title{\textsc{CompAct}: Compressing Retrieved Documents Actively for \\ Question Answering}

\author{Chanwoong Yoon$^1$ \hspace*{0.1cm}
    Taewhoo Lee$^1$ \hspace*{0.1cm}
    Hyeon Hwang$^1$ \hspace*{0.1cm}
    Minbyul Jeong$^{1,2}$\footnotemark[1]\footnotemark[2] \hspace*{0.1cm}
    Jaewoo Kang$^{1,3}$\footnotemark[2] \\
    Korea University$^1$ \\
    Upstage AI$^2$\\
    AIGEN Sciences$^3$ \\
    \{cwyoon99, taewhoo, hyeon-hwang, minbyuljeong,  kangj\}@korea.ac.kr
}

\begin{document}
\maketitle

\footnotetext[1]{This work was done while the author was at Korea University.}
\footnotetext[2]{Corresponding authors.}

\renewcommand{\thefootnote}{\arabic{footnote}}

\begin{abstract}
Retrieval-augmented generation supports language models to strengthen their factual groundings by providing external contexts.
However, language models often face challenges when given extensive information, diminishing their effectiveness in solving questions.
Context compression tackles this issue by filtering out irrelevant information, but current methods still struggle in realistic scenarios where crucial information cannot be captured with a single-step approach.
To overcome this limitation, we introduce \textbf{\textsc{CompAct}}, a novel framework that employs an active strategy to condense extensive documents without losing key information.
Our experiments demonstrate that \textsc{CompAct} brings significant improvements in both performance and compression rate on multi-hop question-answering benchmarks.
\textsc{CompAct} flexibly operates as a cost-efficient plug-in module with various off-the-shelf retrievers or readers, achieving exceptionally high compression rates (47x).\footnote{Code: \href{https://github.com/dmis-lab/CompAct}{https://github.com/dmis-lab/CompAct}.}
\end{abstract}

\section{Introduction}

Retrieval-augmented generation empowers language models to solidify their factual grounding, presenting relevant contexts to answer questions~\citep{khandelwal2019generalization, lewis2020retrieval, karpukhin2020dense, izacard2023atlas}.
While these approaches extend the knowledge scope of language models beyond their inherent capabilities, they also introduce challenges when it comes to handling long contexts~\citep{li2024long, an2024make, qian2024long}. 
First, models often struggle to find key information within extensive contexts, which diminishes their abilities to reference documents~\cite{liu2024lost}.
Also, they often fail to integrate information across multiple documents, which is a common occurrence in real-world scenarios~\citep{cheng2024xrag}.
To this end, there is a growing need for methods that can assist models with handling long contexts.

One way to address these challenges is by compressing contexts into concise forms~\cite{selective-context, pan2024llmlingua2}.
The main goal of compression is to reduce the amount of tokens from the original text without losing core information.
\begin{figure}[t!]
\centering
\includegraphics[width=\columnwidth]{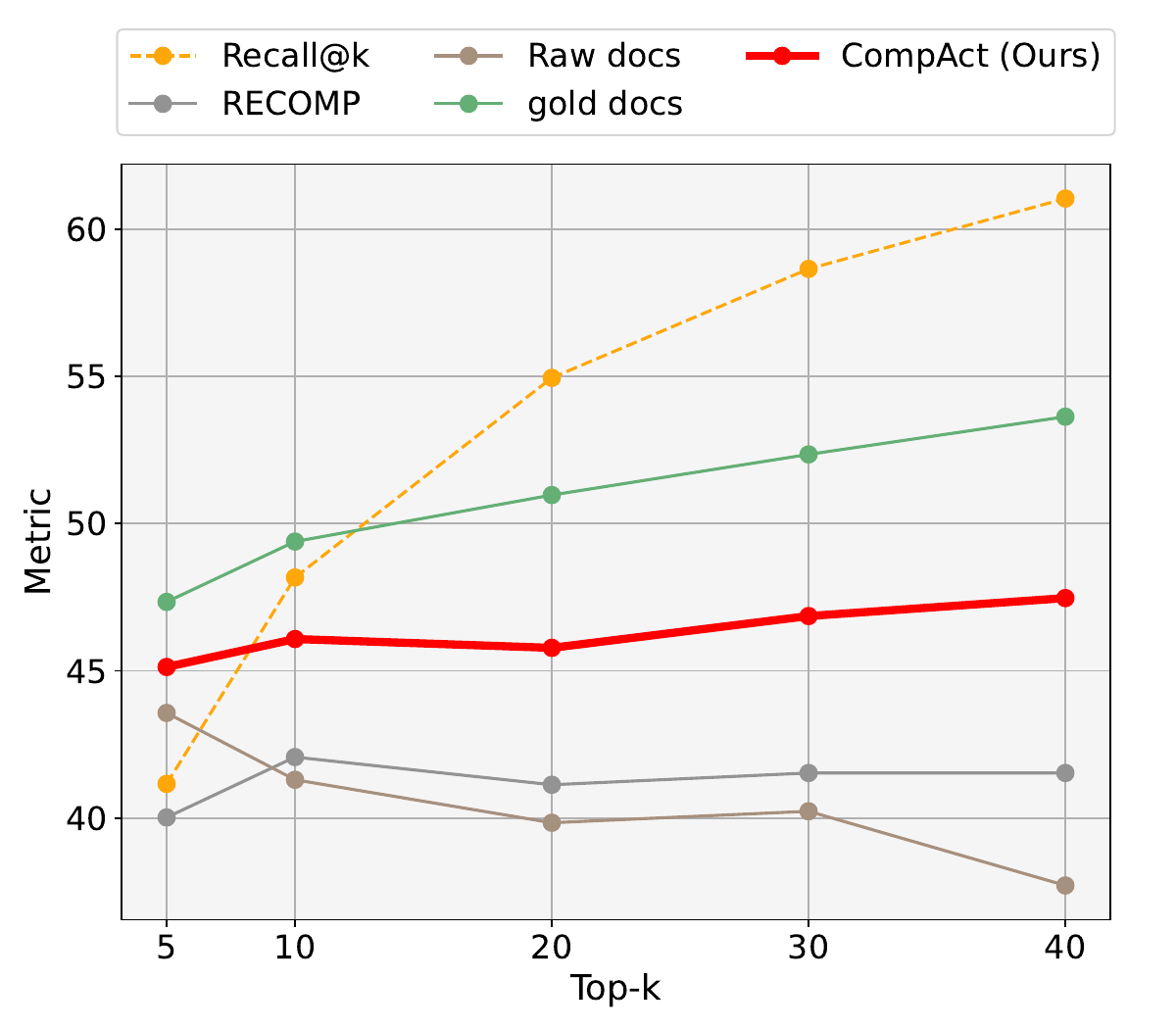}
\caption{
Performance of HotpotQA with different top-$k$ documents, using LLaMA3-8B~\citep{llama3} as the reader.
\textbf{\textsc{CompAct}} shows solid performance improvements that align with those of gold documents. This highlights \textsc{CompAct}'s ability to effectively leverage the benefits of increased top-$k$, unlike other methods that struggle with noisy context.
}
\label{fig:example}
\vspace{-0.3cm}
\end{figure}
However, simply compressing contexts can be suboptimal for QA tasks~\cite{joshi-etal-2017-triviaqa, kwiatkowski2019natural}, where important details may be filtered out during the compression process~\citep{selective-context}.
Maintaining redundant information without compression can harm performance, as it may serve as a distractor that can induce models to generate incorrect responses.
To handle these limitations, query-focused compression emerges as an effective approach, which aims to preserve information relevant to the question~\cite{jiang2023longllmlingua, xu2024recomp, cao2024retaining}.

However, existing query-focused compressors still struggle to take advantage of information located behind lengthy contexts, leaving out potential opportunities for reader models to improve their answers.
In Figure~\ref{fig:example}, the increase in retrieval recall parallel to the number of documents indicates that useful information is still present even in the lower-ranked results.
This demonstrates that these documents should also be covered to fully exploit given information.

\begin{figure*}[t!]
\centering
\includegraphics[width=\textwidth]{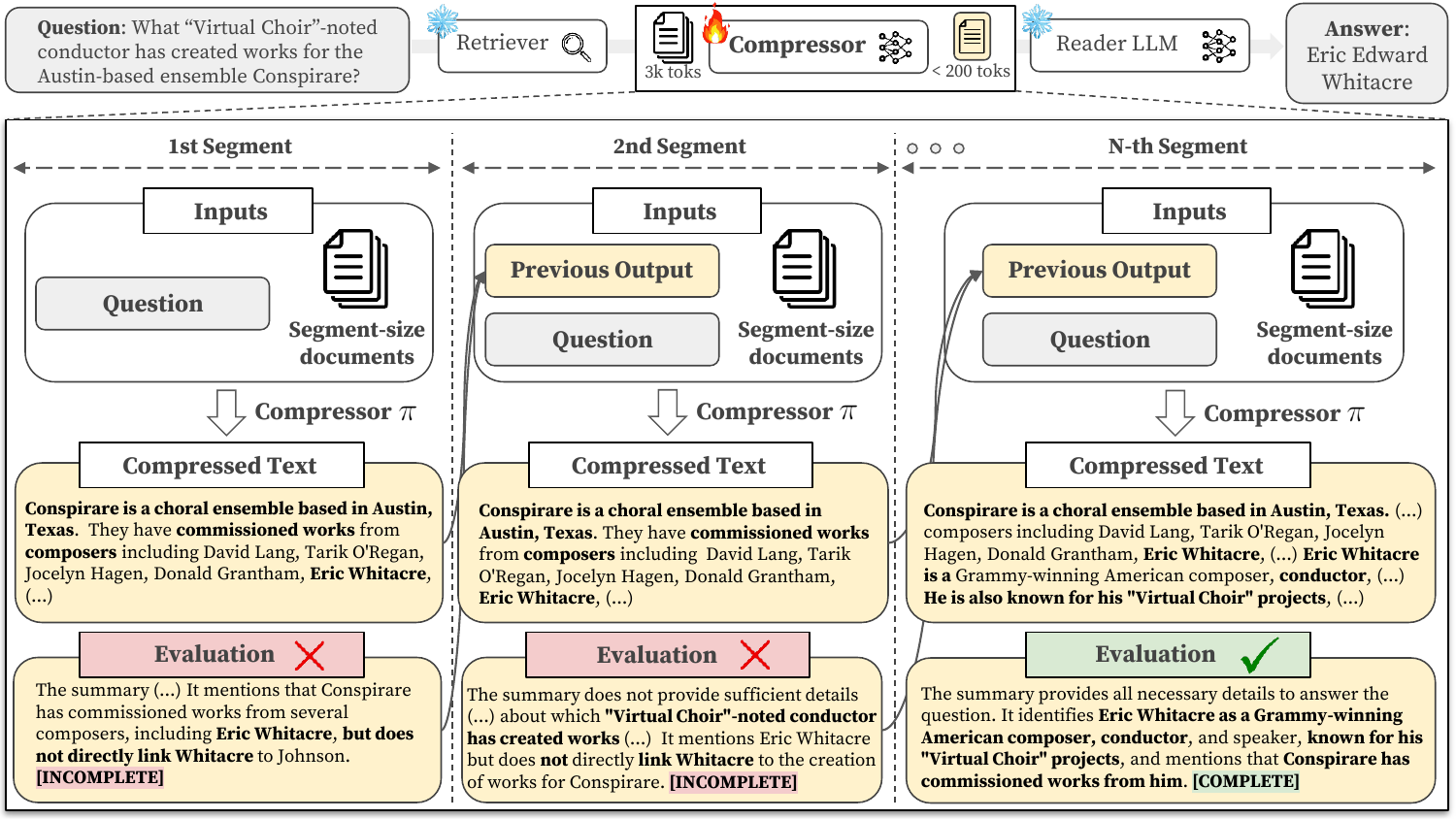}
\caption{
Overall \textsc{CompAct} framework as a plug-in module between the retriever and the reader LLM. 
After splitting retrieved documents into segments, \textsc{CompAct} sequentially compresses these segments into a compacted context. 
By jointly analyzing the previous context with newly provided segments, we actively compress input documents while preserving essential information in the compressed context.
If the segments do not offer complete information to answer the question (1st and 2nd segments), \textsc{CompAct} continues to the next step to acquire new information. 
Once all supporting clues are fully captured ($N$-th segment), the iteration ends.
}
\label{fig:method}
\vspace{-0.3cm}
\end{figure*}

Furthermore, existing methods lack the ability to integrate information across multiple documents, which is required in real-world scenarios~\cite {gutierrez2024hipporag}. 
Figure~\ref{fig:method} depicts an example: the question is \textit{"What `Virtual Choir'-noted conductor has created works for the Austin-based ensemble Conspirare?"}.
To answer this, not only do we need to retrieve information implied within the question, but we should also holistically connect and synthesize information across multiple documents.
In other words, the quality of answers hinges on the ability of models to dynamically integrate information across multiple documents, which is yet to be fully explored in the field of compression.

To this end, we propose \textbf{\textsc{CompAct}}, a novel framework that can address these challenges by using an active strategy to compress extensive documents and retain crucial information.
Our framework has two key components: active compression and early termination. 
During compression, the model actively encapsulates input documents by jointly analyzing previously compressed contexts with newly provided segments. 
This ensures that only the most relevant information (here we refer to the compressed text) to the question is preserved at each step, creating a dense and compact context. 
At each step, the model then decides whether to terminate the compression process. 
This decision is made based on the relevance and completeness of the information gathered to answer the query. 

Our approach offers two distinct advantages. 
First, it effectively captures essential context from long documents by incorporating segments with the previously compressed context. 
This is crucial for complex QA tasks that require in-depth reasoning and synthesis of information.
Second, it condenses large volumes of documents with a high compression rate, without missing essential contexts.
We demonstrate that our framework brings significant improvement in compression rate and end performance in multi-document QA benchmarks.
This highlights the effectiveness of our framework, as it preserves the necessary context without losing critical information.

Our contributions are as follows:
(1) We propose \textbf{\textsc{CompAct}}, a novel framework that employs an active strategy for compressing extensive documents. We address the challenge of handling long contexts by dynamically preserving query-related contexts, focusing on integrating information across documents.
(2) Our framework outperforms existing compressors by a significant margin, achieving a 7.0 (F1) improvement on HotpotQA~\cite{yang2018hotpotqa} with a higher compression rate (47x). Also, it surpasses the performance of long-context large language models in multi-document QA benchmark datasets.
(3) We demonstrate the compatibility of \textsc{CompAct} with various retrievers and readers, underscoring its effectiveness as a plug-in module between retrievers and readers.

\section{Preliminaries}
\subsection{Multi-Document Question Answering}
Multi-document (or multi-hop) question answering (QA) involves the task of answering questions that require gathering information from multiple documents~\citep{yang2018hotpotqa,  ho2020constructing, chen2020hybridqa, trivedi2022musique, mavi2022survey}.
This task is more complicated than single-document QA since it requires models to locate and combine information scattered across multiple sources. 
Even if models can afford lengthy input contexts, they still face challenges in effectively integrating dispersed information from documents.

\subsection{Multi-hop Information-Seeking}
Recent multi-hop information-seeking methods aim to traverse and integrate information across documents by constructing structured maps, such as knowledge graphs or memory trees, over the document context~\citep{kgp, memwalker, readagent}. However, these approaches require an initial building step to create a structured representation of the context. Additionally, they usually navigate their maps to find an optimal traverse path, which demands iterative reasoning by a highly capable model. While we similarly go through the navigation task, we focus on reducing the amount of information the agent has to process, thereby minimizing the computational burden of the reader agent.

\subsection{Context Compression}
Several studies have focused on compressing the inputs of language models to reduce inference cost while preserving core information.
\citet{gisting} introduce gisting, a method that compresses input prompts into shorter transformer activations that can be generalized to unseen prompts.
ICAE~\cite{ge2024incontext} proposes training objectives that compress contexts to be restored as closely as possible to the original.
Selective-Context~\cite{selective-context} and  LLMLingua~\cite{jiang-etal-2023-llmlingua} utilize conditional probabilities of LLMs to assess the importance of information within contexts.
xRAG~\cite{cheng2024xrag} uses modality fusion to embed document representations into language models and achieves high compression rates. 

Additionally, some works have focused on lengthy context inputs.
For example, AutoCompressors~\cite{chevalier2023adapting} transform segments of input context into soft prompts, which are then attached to the next segment as summary vectors.
LongLLMLingua~\cite{jiang2023longllmlingua} select candidates from documents and then perform token-level compression to retain valuable information relevant to a question. 
Concurrent with our work, Chain-of-Agents~\cite{zhang2024chain} has utilized an iterative framework, which enables information aggregation and context reasoning over long-context tasks.
However, our work aims to address a crucial aspect by integrally linking and synthesizing pivotal information between segments while compressing contexts.

\subsection{Task Formulation}

In retrieval-augmented generation, a model $M$ predicts an output $y$ conditioned on an input $x$ and $k$ retrieved passages $D_k = {\{d_1,...,d_k\}}_{i=1}^{k}$.
For the task of question answering, the input $x$ typically consists of a question $q$ along with an instruction.
Thus, $M$ generates an answer $y$ based on $x$ and the retrieved documents $D_k$ as follows: $M(y|x, D_k)$.

To mitigate the costs of $M$ caused by processing a large number of tokens, several approaches have been recently proposed to compress the documents into a shorter context ~\cite{wang2023learning, xu2024recomp}.
Building on these approaches, our goal is described as follows: 
$$
arg\max_{\pi} P_{M}(y \mid C_{\pi}, x)
$$
$$
C_{\pi} = \pi(q, D_{k}) \quad \text{with} \quad l(C_{\pi}) \ll l(D_k) 
$$
where $l$ represents the number of tokens and $\pi$ is a function that compresses documents $D_k$ into a shorter context $C_{\pi}$ based on the question $q$. 
It is important to note that we do not aim to optimize the model $M$ or the retriever.
Instead, our primary focus is on compressing the provided contexts into a concise format to ensure that the essential information is retained for answering the question.

\section{\textsc{CompAct}}
We introduce \textbf{\textsc{CompAct}}, a novel compression framework that actively compresses documents until all necessary evidence for answering a question.
To condense a large amount of information from documents, we devise an iterative architecture where the compressed contexts are updated at each iteration.
In this section, we provide a comprehensive explanation of our framework and detail the data construction process for training our model.

\subsection{Active Compression}
We reconsider compression as sequential updates of contexts based on the previous information.
Figure~\ref{fig:method} clearly shows the concept of our framework.
Given a question and documents $D_k = {\{d_1,...,d_k\}}_{i=1}^{k}$ from a retrieval system, we first group the documents as follows:  
$$
S_{t}=\{d_{(t-1) \times j+1},d_{(t-1) \times j+2},...,d_{(t-1) \times j + j}\}
$$

where $S_{t}$ is a $t$-th segment consisting of $j$ documents, and $j$ represents the predefined number of documents to be compressed at each iteration. For example, $S_1 = \{{d_1, d_2, ..., d_5\}}$ when $j=5$.
We then begin compressing each segment iteratively until it satisfies the end condition. 
It can be formulated as follows:
$$
C_t, E_t = \pi(q, S_t, C_{t-1})
$$

Here, $q$ is a given question to answer. $C_t$ and $E_t$ represent the compressed context and an evaluation at step $t$, respectively. 
$C_t$ is used as part of the input for the next step.
During compression, the model actively integrates information related to the question by jointly analyzing the previously compressed context with a newly provided segment. 
This approach ensures that only the most relevant information is preserved at each step, resulting in a more compact context.
As the output context is designed to preserve query-related information, it serves as a comprehensive memory of all iterations up to the current step.
We describe an example in Table \ref{table:samples}. 

\subsection{Early Termination}
To ensure that the iteration does not continue unnecessarily once enough information is obtained, we introduce a specific end condition for early termination.
We implement this by including an evaluation $E$ in the generation process to decide the endpoint.
The evaluation $E$ consists of a rationale and a condition token  ([COMPLETE] or [INCOMPLETE]).
The purpose of $E$ is to assess whether an input segment $S_t$, combined with the previous context $C_{t-1}$, provides sufficient details to answer the question.
If the token indicates that the provided context is sufficient, the iteration terminates; otherwise, we continue to gather missing information until all details are fully obtained.

This early termination offers three primary benefits. 
First, it prevents redundant contexts from entering the compressed contexts or acting as a distraction.
Second, it avoids meaningless iterations, thereby drastically lowering the computational burden that may stem from iterative processing steps.
Third, it dynamically adjusts to the complexity of the question and the information density of the documents. 
This flexibility enables our \textsc{CompAct} framework to be both effective and efficient across a wide range of scenarios, from simple questions to more complex, multi-hop questions that require extensive context integration.

\begin{figure}[t!]
\centering
\includegraphics[width=.98\columnwidth]{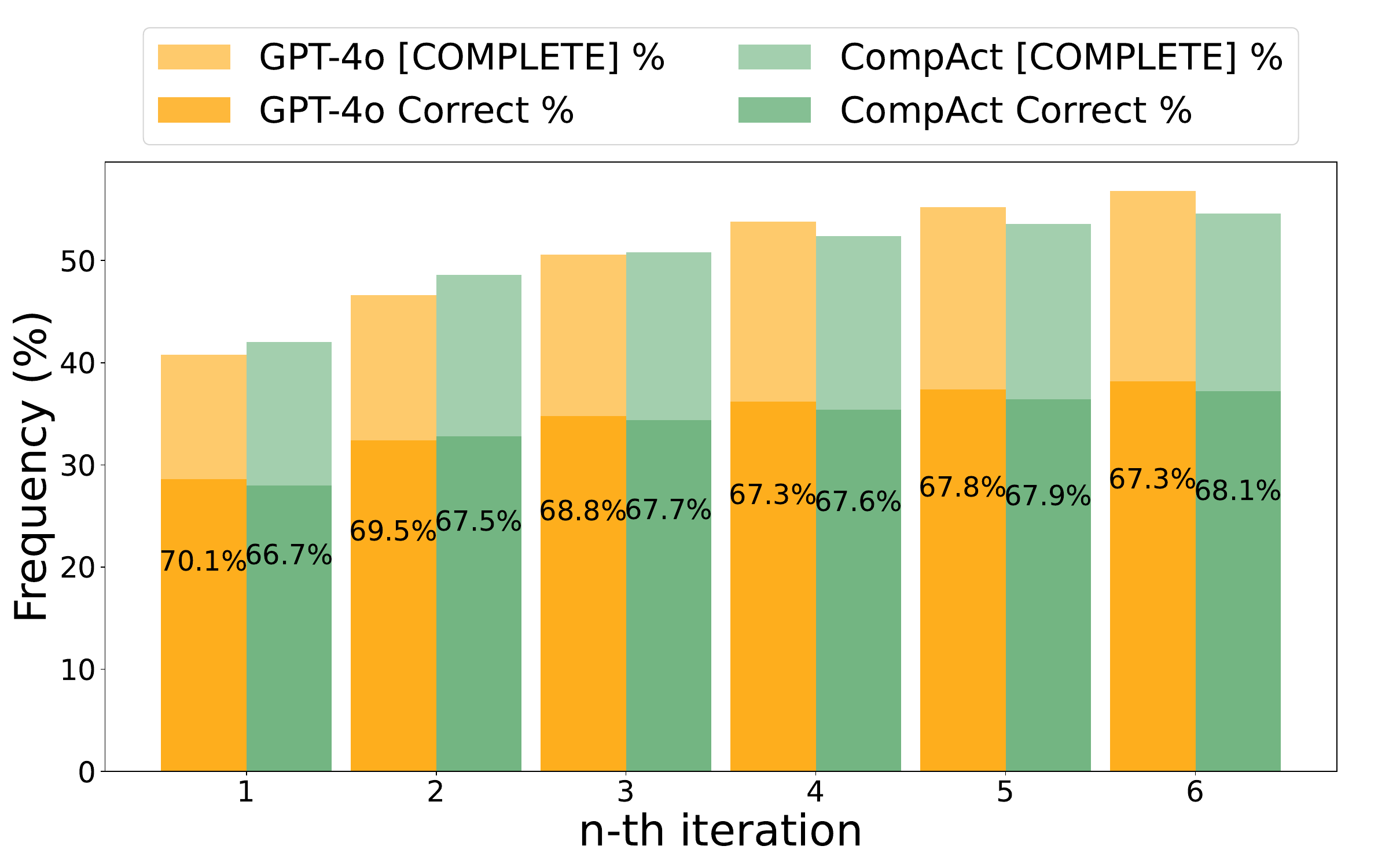}
\caption{
Distribution of iteration points where models determine the compressed contexts to be complete. 
The frequencies of completeness are accumulated over iterations. 
We compare the distribution between GPT-4o (Yellow) and \textsc{CompAct} (Green).
We also measure the percentage of correctness at each iteration, using an F1 score of 0.4 as the threshold for correction.
}
\label{fig:completeness}
\vspace{-0.3cm}
\end{figure}

\subsection{Dataset Construction}
We compress documents into a query-related context while concurrently determining the endpoint of the iterations. 
To cultivate this capability, we instruct a superior LLM to follow a three-step process.
We provide the prompt in Table~\ref{table:prompt_dataset_construction}.

\paragraph{Sentence-Level Selection.}
We begin by asking the LLM to analyze sentences, particularly focusing on relevant clues that may help answer the question.
If certain sentences provide relevant information or implicitly clarify ambiguous points within the question, the LLM is instructed to generate these sentences from the provided documents.

\paragraph{Query-focused Compression.} 
We then generate a compressed text of the selected sentences based on the question.
We explicitly restrict the LLM from making assumptions or attempting to conclude without supporting evidence, as follows: \textit{"DO NOT make assumptions or attempt to answer the question; your job is to summarize only."}.
This restriction is crucial because our main objective here is to condense relevant information from the provided documents, instead of directly answering the questions. 
Skipping the logical steps required to answer the question, as if relying on parametric knowledge, can harm compression performance by increasing the likelihood of missing essential information.

\begin{table}[t!]
\centering
{\resizebox{0.48\textwidth}{!}{
\begin{tabular}{l cc cc c}
\toprule
\multirow{3}{*}{\textbf{Dataset}} & \multicolumn{2}{ c}{\textbf{[COMPLETE]}} & \multicolumn{2}{ c}{\textbf{[INCOMPLETE]}} & \multirow{3}{*}{\textbf{Total}} \\ \cmidrule{2-5}
 & \multicolumn{1}{c}{\textbf{first}} & \multicolumn{1}{c}{\textbf{subsequent}} & \multicolumn{1}{c}{\textbf{first}} & \multicolumn{1}{c}{\textbf{subsequent}} &        \\ \midrule
HotpotQA & \multicolumn{1}{c}{7.2K}    & 7.2K    & \multicolumn{1}{c}{7.2K}    & 7.2K    & 28.8K        \\ 
\bottomrule
\end{tabular}}}{}
\caption{
Statistics of our generated dataset. 
We categorize it into four cases: [COMPLETE] and [INCOMPLTE], each further split based on whether it is the first or subsequent iteration.
}
\label{table:dataset}
\vspace{-0.3cm}
\end{table}

\paragraph{Determining the Early Termination.}
We also prompt the LLM to evaluate its own compressed contexts based solely on the provided information, without any additional background context.
We direct the LLM to generate a condition token (e.g., [COMPLETE] or [INCOMPLETE]) along with the rationale for its judgment.

Overall, we construct a synthetic dataset by instructing the LLM based on the three-step processes described above. Table~\ref{table:dataset} shows the dataset statistics.
We conduct data construction from two scenarios: realistic and distractor.
In realistic scenarios, provided documents are the results of a retrieval system.
However, due to the retriever's limited performance, gold documents rarely appear, which can hinder the collection of cases with early termination.
This results in a scarcity of cases in the dataset where the iteration is terminated early (i.e. [COMPLETE] at a subsequent iteration). 
To address this issue, we collect data from distractor scenarios which include predefined documents that contain all supporting facts needed to answer the question.
After filtering the collected datasets from both scenarios, we build a training dataset consisting of 28k instances categorized into four distinct groups.
\section{Experiment}
\begin{table*}[t!]
\centering
\resizebox{1.0\textwidth}{!}{
\begin{tabular}{l ccc ccc ccc ccc ccc}
\toprule
  {\multirow{2}{*}{\textbf{Methods}}} &    \multicolumn{3}{c}{\textbf{HotpotQA}} & \multicolumn{3}{c}{\textbf{MuSiQue}} & \multicolumn{3}{c}{\textbf{2WikiMQA}} & \multicolumn{3}{c}{\textbf{NQ}} & \multicolumn{3}{c}{\textbf{TriviaQA}} \\ \cmidrule{2-16} 
  {}     &     {\textbf{Comp.}} &   {\textbf{EM}}   &   {\textbf{F1}}   &   {\textbf{Comp.}} &   {\textbf{EM}}   &   {\textbf{F1}}  &    {\textbf{Comp.}} &   {\textbf{EM}}   &   {\textbf{F1}} &    {\textbf{Comp.}} &   {\textbf{EM}}   &   {\textbf{F1}} &    {\textbf{Comp.}} &   {\textbf{EM}}   &   {\textbf{F1}}  \\ \midrule
\multicolumn{1}{l}{Oracle}     &      10.8x     &  39.9 & 51.2  & 10.3x & 14.2 &  23.6 & 11.0x & 37.4 & 43.2 &  -  & -  & - & - & - & - \\

\multicolumn{1}{l}{Raw Document}     &      {1x}  & {29.4}  &  {40.3} & {1x}  & {6.5} & {15.6} & {1x}  & 25.4  & 31.2 &  {1x}  & {39.0} & {51.3}  &  {1x} & {68.9}  & 77.1 \\ 
\midrule
\multicolumn{16}{c}{\textit{Long-Context LLM}} \\ \midrule
\multicolumn{1}{l}{InternLM2-chat-7B}    & {1x}  & {8.0} & {20.3} & {1x} & {1.0} & {6.8} &  {1x} & {9.3} & {19.5} &  {1x}  & {7.6}   & {22.6} & {1x} & {12.1} & {31.5} \\
\multicolumn{1}{l}{Mistral-7B-Instruct-v0.2}      & {1x}  & {9.5} &   {22.6} & {1x}  & {1.0}  & {7.9} & {1x}  & {1.2}  & {15.4} & {1x} & {4.3} & {20.9}  & {1x} & {35.3}   & {50.4} \\
\multicolumn{1}{l}{FILM-7B}   & {1x}  & {32.4} &   {43.7} & {1x} & {6.9} & {15.7} & {1x}  & {26.4}  &  {31.7} &  {1x}  & {38.2} & {50.8} & {1x} & {62.7} & {71.7} \\
\multicolumn{1}{l}{Phi-3-medium-128k-instruct}     & {1x}  &  {22.3} & {34.7} & {1x} & {5.8}  & {14.6} & {1x}  & {24.8} & {31.0} & {1x} & {29.1} & {42.2} & {1x} & {61.0} & {70.6} \\
\multicolumn{1}{l}{Yi-9B-200k} & {1x}  &  {28.6} & {39.4} & {1x} & {6.8} & {15.1} & {1x}  & {25.0} & {30.4} & {1x} & {33.9} & {45.2} & {1x} & {62.8} & {71.3} \\
\multicolumn{1}{l}{Phi-3.5-mini-instruct} & {1x}  &  {21.6} & {33.0} & {1x} & {4.8} & {12.7} & {1x}  & {21.0} & {26.8} & {1x} & {29.3} & {41.2} & {1x} & {58.2} & {67.9} \\
\multicolumn{1}{l}{Llama-3.1-8B-Instruct} & {1x}  &  {31.4} & {42.9} & {1x} & {6.2} & {14.6} & {1x}  & {30.2} & {36.0} & {1x} & {35.8} & {49.3} & {1x} & {63.6} & {74.2} \\
\multicolumn{1}{l}{GPT-3.5-turbo}     & {1x}  &  {32.8} &   {43.8} &  {1x}  & {7.3}  & {16.1} & {1x}  & {28.6}  & {33.9} & {1x} & \textbf{40.8} & \textbf{54.6} & {1x} & \textbf{69.9} & \textbf{77.4} \\
\midrule
\multicolumn{16}{c}{\textit{Compressor}} \\ \midrule
\multicolumn{1}{l}{AutoCompressors}   &  {35.4x} & {18.4} & {28.4}  & {34.7x}  & {3.9} & {11.9} & {36.2x}  & {19.0}  &  {24.5}   &  {34.4x}  & {17.3} & {31.8} & {34.5x} & {55.3} & {64.3} \\ 
\multicolumn{1}{l}{LongLLMLingua}      &   {3.4x}     &   {25.6}   &   {35.3}  & {3.4x}  & {4.8} & {13.5} & {3.6x}  & {27.9}  & {32.9} & {3.5x} & {27.7} & {40.6}   &   {3.3x} & {64.0} & {70.8} \\ 
\multicolumn{1}{l}{RECOMP (extractive)}     &   {34.3x}  &   {29.7}          &   {39.9}      &   {32.7x}   &   {6.7}   &   {15.7}    & {35.9x}      &   {29.9}   &   {34.9}  & {32.7x}  & {34.6}  & {45.1}  &  {39.2x} & {67.6}  & {74.1}  \\ 
\multicolumn{1}{l}{\textsc{CompAct} (Ours)} &  \textbf{47.6x} &  \textbf{35.5}  &  \textbf{46.9}  &  \textbf{37.2x} &  \textbf{8.7} &  \textbf{18.1}  &  \textbf{51.2x} & \textbf{31.0} & \textbf{37.1} &  \textbf{48.5x} & 38.4 &  50.0 &  \textbf{49.4x} & 65.4 & 74.9 \\ 
\bottomrule
\end{tabular}}
\caption{Main results. We set the reader as LLaMA3-8b~\citep{llama3}. 
We retrieve top-30 documents. 
We use three multi-document (HotpotQA, MuSiQue, and 2WikiMQA) and two single-document (NQ and TriviaQA) question-answering datasets.
Since our training datasets consist of a subset of HotpotQA, we perform zero-shot evaluation on the rest of the datasets.
Comp. refers to the compression rate which is denoted as follows: $\text{compression rate}=\frac{\text{\# of tokens in retrieved documents}}{\text{\# of tokens in compressed text}}$.
}
\label{table:main}
\vspace{-0.3cm}
\end{table*}

\subsection{Experimental Setup}
\paragraph{Dataset Construction}
We use the GPT-4o~\cite{gpt-4o} API (2024-05-13) as the LLM to collect our dataset.
We use only a subset of HotpotQA~\cite{yang2018hotpotqa} training set for data collection.
To retrieve documents, we use Contriever~\cite{izacard2022unsupervised}, fine-tuned on MS-MARCO~\cite{bajaj2016ms}, as our retrieval system on the 2018 Wikipedia corpus~\cite{karpukhin-etal-2020-dense}.
We set the default number of documents per segment $j$ to 5 and top-$k$ to 30, allowing for a maximum of 6 iterations per query. 
To prevent lengthy API responses, the maximum number of generated tokens is limited to 700.

\paragraph{Training \& Inference}
We perform supervised fine-tuning to train our model using the collected dataset.
Without using specific labels or methods for particular iterations, we focus on training the model to effectively update the previous context based on the question and given documents at each step.
We use instruction-tuned Mistral-7B~\cite{jiang2023mistral} as our backbone base model. 
At inference, we process the same number of segments and inputs as training. 
Further information is provided in the Appendix~\ref{app:details}.

\subsection{Datasets} 
We evaluate \textsc{CompAct} on both single-document and multi-document question-answering (QA) datasets.
For single-document QA, we use Natural Question (NQ)~\cite{kwiatkowski2019natural} and TriviaQA (TQA)~\cite{joshi-etal-2017-triviaqa}.
For multi-document QA, we evaluate on HotpotQA~\cite{yang2018hotpotqa}, MuSiQue~\cite{trivedi2022musique}, and 2WikiMultiHopQA~\cite{ho-etal-2020-constructing}.
The evaluation is conducted on the dev set of each dataset, except for TriviaQA, which is evaluated on the test set.
As mentioned, we construct the training data using only a subset of HotpotQA. Therefore, we conducted zero-shot evaluation on the other datasets without accessing their training set.

\subsection{Baselines}
In Table~\ref{table:main}, we compare \textsc{CompAct} against several baseline methods. 
To ensure a fair comparison, we feed compressed contexts from each baseline to the same reader model, LLaMA3-8B~\citep{llama3}. We consider the following baselines.
(1) \textit{Oracle.} We provide the reader with documents that contain the answers to the questions. If such documents are not available, we include five documents as a default.
(2) \textit{Raw Document.} We simply concatenate the top-$k$ retrieved documents.
(3) \textit{Long-Context LLM.} As these LLMs are designed to handle large inputs, they align with our objective of managing extensive contexts, making them suitable for our baselines. 
We use InternLM2-chat-7B~\citep{cai2024internlm2}, Mistral-7B-Instruct-v0.2~\citep{jiang2023mistral}, FILM-7B~\citep{an2024make}, Phi-3-medium-128k-instruct and Phi-3.5-mini-instruct~\citep{phi}, Yi-9B-200k~\citep{yi}, Llama-3.1-8B-instruct~\citep{llama3}, and GPT-3.5-turbo-0125~\citep{openai2023a}.
(4) \textit{Compressor.} We compare \textsc{CompAct} with three compression-based methods: AutoCompressors~\cite{chevalier2023adapting}, RECOMP~\cite{xu2024recomp}, and LongLLMLingua~\cite{jiang2023longllmlingua}.
We provide the detailed descriptions of the baselines in Appendix~\ref{app:baselines}.

\subsection{Results}
We assess the performance of \textsc{CompAct} using three metrics: Compression rate (Comp.), Exact Match (EM), and F1 score. 
Overall, \textsc{CompAct} exhibits strong performance across all QA benchmark datasets, achieving the highest compression rate among all baselines.
Specifically, it surpasses other compression-based methods in all three metrics, demonstrating its strong ability to compress abundant information ($\sim$3k tokens) efficiently.

\textsc{CompAct} falls short of the performance of GPT-3.5-turbo in single-document QA (NQ and TriviaQA), which may be due to our model being trained exclusively on a subset of HotpotQA. 
Even with this constraint, our framework outperforms existing compressors and achieves competitive performance with long-context LLMs.
Plus, it represents entire contexts using significantly fewer tokens, highlighting its efficiency in providing compact representations. 
Moreover, in multi-document QA, \textsc{CompAct} achieves superior performance compared to other baselines. This underscores the persistent challenge of integrating information across multiple documents and emphasizes how \textsc{CompAct} excels at such tasks.
\section{Analysis}

We investigate ways to facilitate the use of \textsc{CompAct} as a plug-in module that collaborates with diverse retrievers and readers (Section~\ref{ana:plug-in}).
We conduct an ablation study to assess the impact of components on performance (Section~\ref{ana:component}) and examine the cost-effectiveness of our framework using proprietary black-box models (Section~\ref{ana:cost}). 
Finally, we discuss computational efficiency involved in our framework (Section~\ref{ana:compute}).

\subsection{Compressor as a Plug-in Module}
\label{ana:plug-in}
In Figure~\ref{fig:method}, we illustrate the compressor as a plug-in module, highlighting that retrievers and readers can be easily replaced by other models.
We investigate if \textsc{CompAct} can flexibly compress context provided by diverse retrievers, while preserving useful information regardless of various readers.

\paragraph{Generalizability across Retrievers.}
\begin{figure}[t!]
\centering
\includegraphics[width=.9\columnwidth]
{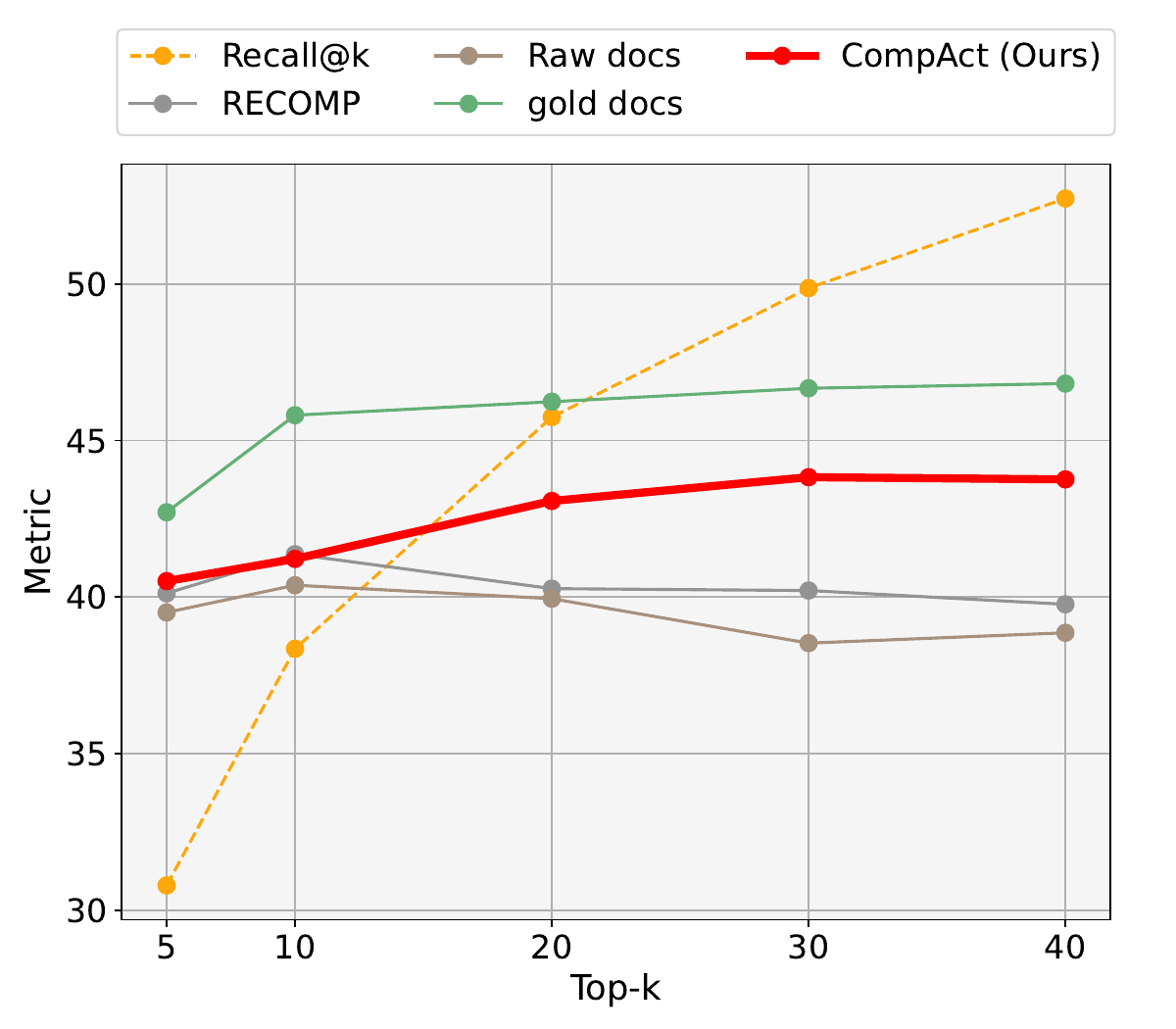}
\includegraphics[width=.9\columnwidth]
{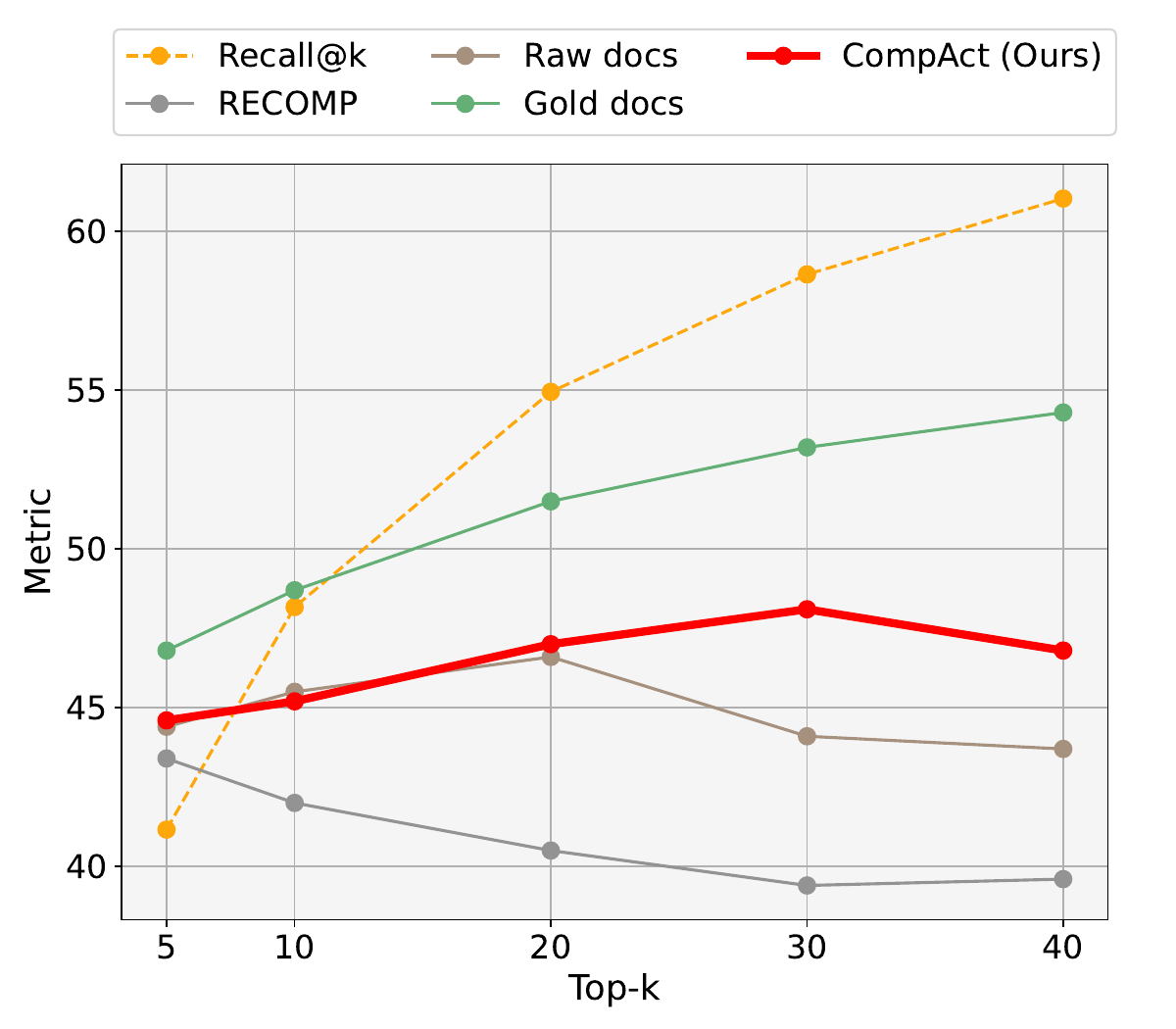}
\caption{
Performance of HotpotQA with different top-$k$ documents, using Contriever (upper) as the retriever and GPT-3.5-Turbo (lower) as the reader.
}
\label{fig:3.5_reader}
\vspace{-0.3cm}
\end{figure}
In Figure~\ref{fig:3.5_reader} and \ref{fig:bm25}, we use Contriever~\citep{izacard2022unsupervised} and BM25~\citep{robertson2009probabilistic}, two of the most well-known retrievers, to replace source documents.
We evaluate our framework with 500 random samples from the HotpotQA~\citep{yang2018hotpotqa} dev set, using different top-$k$.
We compare our results with several baselines: gold documents (oracle), raw documents, and RECOMP~\citep{xu2024recomp}.

With the Contriever setup, where the retriever often fails to locate relevant documents at high-ranking positions, increasing the top-$k$ leads to more distinct performance improvements. 
This shows that our framework effectively captures and utilizes valuable information from lower-ranked documents.
Additionally, in the BM25 setup, \textsc{CompAct} shows consistent performance while retrieving up to top-40 documents. Notably, our framework achieves a similar saturated performance trend to the gold documents setup, indicating its competence in filtering noisy contexts.
In both setups, \textsc{CompAct} achieves significantly higher performance compared to other baselines.
As we intended, these observations demonstrate that \textsc{CompAct} shows robustness across various retriever setups.

\paragraph{Generalizability across Readers.}
\begin{table}[t!]
\centering
{\resizebox{0.48\textwidth}{!}{
\begin{tabular}{l cc cc cc}
\toprule
\multicolumn{1}{c}{\multirow{2}{*}{\textbf{Components}}} & \multicolumn{2}{c}{\textbf{HotpotQA}}            & \multicolumn{2}{c}{\textbf{MuSiQue}}         & \multicolumn{2}{c}{\textbf{2WikiMQA}} \\ \cmidrule{2-7} 
\multicolumn{1}{c}{}       & \multicolumn{1}{c}{\textbf{Comp.}}  & \textbf{F1}   & \multicolumn{1}{c}{\textbf{Comp.}}  &  \textbf{F1} & \multicolumn{1}{c}{\textbf{Comp.}}  & \textbf{F1} \\ \midrule
\multicolumn{7}{c}{LLaMA3-8B}  \\
\midrule
Rationale. & 130.8x & 41.6 & 120.0x & 15.9 & 141.3x &  32.3  \\ 
CT & 47.5x & \textbf{48.3} & 36.5x  & \textbf{19.1} & 52.2x & \textbf{36.2}   \\ 
CT + Rationale & 33.6x & 47.3 & 27.1x & 19.0 & 36.4x &  35.6   \\  
\midrule
\multicolumn{7}{c}{LLaMA2-13B}  \\ 
\midrule
Rationale. & 141.8x & 41.8 & 129.2x  & 16.9 & 152.4x &  30.8  \\ 
CT & 48.1x & \textbf{48.5} & 37.0x  & \textbf{18.6} & 52.7x &  \textbf{35.6}  \\ 
CT + Rationale. & 34.6x & 47.3 & 28.0x  & \textbf{18.6} & 37.4x &  34.2   \\ 
\midrule
\multicolumn{7}{c}{GPT-3.5-Turbo}  \\ 
\midrule
Rationale. & 135.2x & 38.0 & 123.5x & 13.8 & 146.2x &  24.0  \\ 
CT & 48.1x & \textbf{49.2} & 37.0x & \textbf{20.9} & 53.0x &  34.0  \\ 
CT + Rationale. & 33.9x & 47.0 & 27.4x & 18.5 & 36.7x & \textbf{36.5}   \\  
\bottomrule
\end{tabular}}}{}
\caption{Results of each component effectiveness. CT refers to the compressed text.}
\label{tab:component}
\vspace{-0.3cm}
\end{table}
We look into whether \textsc{CompAct} truly provides generalized compressed texts suitable for diverse readers.
To this end, we assess the quality of our compressed texts using diverse reader LLMs: GPT-3.5-Turbo~\citep{openai2023a}, LLaMA2-13B~\citep{touvron2023llama}, and LLaMA3-8b~\citep{llama3}.
Figure~\ref{fig:3.5_reader} presents the results of using GPT-3.5-Turbo as a reader, while figure~\ref{fig:reader_llama} includes the results for LLaMA2-13B and LLaMA3-8B.

Our results show that \textsc{CompAct} sufficiently delivers high-quality compressed texts applicable to different readers.
Also, we prove its effectiveness on the top-$k$ documents with high $k$.
In Figure~\ref{fig:3.5_reader}, there is little difference in performance up to the top-20 between the raw documents setup and ours.
We hypothesize this is attributed to the strong performance of the reader, GPT-3.5-Turbo, in processing moderate length of contexts.
However, at the top-30 and top-40 documents, performance degradation occurs as more documents are included, reflecting the difficulty of handling lengthy documents with increased noisy information.
In contrast, \textsc{CompAct} exhibits marginal performance degradation even with a higher number of documents.

Furthermore, \textsc{CompAct} achieves a high compression rate above 40x, which significantly reduces the number of input tokens, making it highly cost-effective for API operations.
This efficiency, combined with its ability to maintain performance across diverse readers, underscores the superior capability of \textsc{CompAct}.

\subsection{Component Effectiveness}
\label{ana:component}
\textsc{CompAct} actively compresses source documents by generating an intermediate compressed text (CT) with termination evaluation for each iteration.
The evaluation consists of two components: a rationale explaining the reasons for termination and a condition token to decide the termination.
To understand how each component affects end performance, we conduct an ablation study of components as shown in Table~\ref{tab:component}.
we use 500 random samples from each dataset.
When only the rationale is provided, the compression rate increases dramatically, but the end performance (EM \& F1) significantly drops (Row 1).
Conversely, when we only provide compressed text, we achieve the highest performance with most readers. 
However, when adding the rationale with the compressed text (CT + Rationale), there are no clear benefits; in most cases, performance declines.
We hypothesize that some judgments in the rationale distract the readers from generating an answer purely from the compressed context. 
This could act as a negative shortcut in the answering process, resulting in decreased performance.

\begin{table}[t!]
\centering
{\resizebox{0.48\textwidth}{!}{
\begin{tabular}{l cc cc cc cc}
\toprule
\multicolumn{1}{c}{\multirow{2}{*}{\textbf{Model}}} & \multicolumn{2}{c}{\textbf{Raw}} & \multicolumn{2}{c}{\textbf{\textsc{RECOMP}}} & \multicolumn{2}{c}{\textbf{\textsc{Lingua*}}} & \multicolumn{2}{c}{\textbf{\textsc{CompAct}}}  \\ \cmidrule{2-9} 
\multicolumn{1}{c}{}       & \multicolumn{1}{c}{\textbf{Cost}}  & \textbf{F1} & \multicolumn{1}{c}{\textbf{Cost}}  & \textbf{F1} & \multicolumn{1}{c}{\textbf{Cost}}  & \textbf{F1}   & \multicolumn{1}{c}{\textbf{Cost}}  &  \textbf{F1} \\
\midrule
GPT-3.5-Turbo & 1.09 & 44.5 & \textbf{0.04} & 40.1 & 0.33 & 38.4 & \textbf{0.04} & \textbf{49.2} \\
GPT-4o & 10.75 & 55.8 & 0.43 & 48.1 & 3.31 & 47.6 & \textbf{0.28} & \textbf{56.0} \\
Claude-3.5 & 6.45 & 36.0 & 0.26 & 37.0 & 1.99 & 30.2 & \textbf{0.17} & \textbf{42.2} \\
Gemini-1.5-pro & 7.54 & \textbf{52.0} & 0.31 & 41.7 & 2.36 & 40.1 & \textbf{0.20} & 44.8 \\
\bottomrule
\end{tabular}}}{}
\caption{API cost of 500 samples from a HotpotQA dev set. \textsc{Lingua*} refers to LongLLMLingua. We assess the inference cost (USD) of each method when employing proprietary models as readers.}
\vspace{-0.3cm}
\label{table:cost_efficiency}
\end{table}
\subsection{Cost Efficiency}
\label{ana:cost}
To evaluate the cost-saving benefits, we employ four proprietary models as readers: GPT-3.5-Turbo~\cite{openai2023a}, GPT-4o~\cite{gpt-4o}, Claude-3.5-sonnet~\cite{claude-3.5-sonnet}, and Gemini-1.5-pro~\cite{gemini-1.5-pro}.
In Table~\ref{table:cost_efficiency}, we show that our framework achieves superior performance at the lowest cost compared to other baselines.
Surprisingly, \textsc{CompAct} achieves competitive performance to the raw document setups with superior models known to possess exceptional long-context understanding ability. This indicates \textsc{CompAct}'s high-level expertise in compressing contexts.

\begin{table}[t!]
\centering
\resizebox{0.48\textwidth}{!}{
\begin{tabular}{lcccc}
\toprule
\textbf{Dataset} & \multicolumn{2}{c}{\textbf{TFLOPs}} & \multicolumn{2}{c}{\textbf{F1}} \\
\cmidrule(lr){2-3} \cmidrule(lr){4-5}
 & \textbf{Raw} & \textbf{CompAct} & \textbf{Raw} & \textbf{CompAct} \\
\midrule
HotpotQA  & \textbf{34.1} & 35.8 & 40.0 & \textbf{48.3} \\
MusiQue   & \textbf{33.6} & 49.3 & 16.2 & \textbf{19.0} \\
2WikiMQA  & \textbf{35.9} & 42.4 & 29.5 & \textbf{37.2} \\
NQ        & 32.9 & \textbf{26.7} & 52.9 & \textbf{53.8} \\
TQA       & 33.5 & \textbf{24.6} & \textbf{78.5} & 77.3 \\
\bottomrule
\end{tabular}
}
\caption{Average TeraFLOPs (TFLOPs) and F1 scores. TFLOPs are normalized by the number of instances. We utilize LLaMA3-8B as a reader with top-30 documents and employ DeepSpeed FlopsProfiler~\cite{deepspeed} for measurement.}
\label{table:computational_efficiency}
\end{table}

\subsection{Computational Efficiency}
\label{ana:compute}
While \textsc{CompAct} offers a significant cost-saving advantage, we also consider a potential increase in computation due to the active strategy employed by our framework.
To assess this, we measure the Floating Point Operations per Second (FLOPs) of our framework in comparison to the baseline raw context setup, as shown in Table~\ref{table:computational_efficiency}.

We reveal that \textsc{CompAct} consistently demonstrates higher performance than the raw context setup in multi-hop QA tasks (HotpotQA, MusiQue, 2WikiMQA). 
Specifically, on HotpotQA, it achieves a large increase in F1 score while maintaining comparable computational costs. 
Although MusiQue and 2WikiMQA exhibit higher costs-primarily due to the increased iterations required to identify supporting documents in low recall scenarios-the performance gains are substantial.
Conversely, for single-hop QA tasks (NQ, TQA), our framework achieves competitive performance with significantly reduced costs, demonstrating the saving effect of early termination. This highlights how the active strategy allows us to dynamically adjust the computational costs allocated to each instance, making our framework flexible for questions with diverse levels of complexity and varying information requirements.

\section{Conclusion}
We introduce \textbf{\textsc{CompAct}}, a novel framework that employs an active strategy to compress extensive retrieved documents.
Our framework effectively captures pivotal information from a large number of documents by dynamically retaining essential contexts and incorporating information.
We demonstrate that \textsc{CompAct} significantly outperforms existing compressors, showing a large performance gap with a higher compression rate in multi-document question-answering benchmarks.
Furthermore, it serves as a convenient plug-in module that can seamlessly collaborate with various off-the-shelf retrievers and readers while providing cost-saving benefits.

\section*{Limitations}
We acknowledge that \textsc{CompAct} has a longer inference time when processing retrieved documents, compared to other compressors.
Given that our framework contributes to addressing complex question types, which is pioneering in the field of compression, we believe that future research can build upon  \textsc{CompAct} to further improve these issues.

Additionally, even a strong proprietary model like GPT-4o can make mistakes when determining the completeness of given contexts. 
There may still be error cases in our data construction process, although we attempt to address this issue by filtering them out.

Lastly, we only use Mistral-7B-Instruct-v0.2 as our base model due to resource limitations.
Verifying whether \textsc{CompAct} works well across a range of model sizes, both smaller (< 7B) and larger (> 7B), could lead to interesting findings.

\section*{Ethics Statement}
Our training process can incur significant environmental costs due to its computationally intensive nature. 
To mitigate this, we fine-tune a single Mistral model to minimize computational expenses.
Furthermore, a potential risk of this work is that the generated dataset may contain biases from API calls, such as stereotypes related to race and gender.
To our knowledge, there haven't been significant issues reported when creating question-answering datasets.
However, it would be beneficial to apply methods that robustly train or validate against such concerns.
\section*{Acknowledgments}
This work was supported in part by the National Research Foundation of Korea [NRF2023R1A2C3004176], the Ministry of Health \& Welfare, Republic of Korea [HR20C002103], the Ministry of Science and ICT (MSIT) [RS-2023-00220195], the ICT Creative Consilience program through the Institute of Information \& Communications Technology Planning \& Evaluation (IITP) grant funded by the MSIT [IITP-2024-2020-0-01819], and the National Research Foundation(NRF), Korea, under project BK21 FOUR.


\appendix
\label{sec:appendix}

\section{Practicality of Compressing Contexts}
\label{appendix:practicality}

To ensure the practicality of providing context with fewer tokens, we present an additional point to reinforce the necessity of our research.
In table \ref{table:huggingface_model}, we investigate the maximum input length of language models with over 1 million downloads on Huggingface\footnote{\href{https://huggingface.co/Models}{https://huggingface.co/Models}}.
We find that 77.5\% of these models can only afford inputs of 512 tokens or fewer.
Despite ongoing research trends on LLMs capable of handling long texts, it is evident that many users still frequently employ models with smaller token inputs.
Considering the current state, \textsc{CompAct} offers substantial benefits to models with smaller input lengths by allowing them to access more information, effectively acting as a bridge.
\begin{table}[h]
\centering
{\resizebox{0.48\textwidth}{!}{
\begin{tabular}{cc}
\toprule
\textbf{Sequence Length} & \textbf{Language Models (\%)} \\ \midrule
128                                                       & 14.7                 \\ 
512                                                       & 62.8                 \\ 
$\geq$ 1024                                                & 22.5                 \\ \bottomrule
\end{tabular}}}{}
\caption{Huggingface Models Statistics. 77.5\% of models cannot receive at least top-5 documents as input. We select frequently-used models downloaded at least 1M in \href{https://huggingface.co/Models}{https://huggingface.co/Models}.}
\label{table:huggingface_model}
\end{table}

\section{Additional Comparison}
To further evaluate against additional compression methods, we conduct supplementary experiments following the setup from~\citet{cao2024retaining}. The results show that our framework outperforms existing methods on both TriviaQA and HotpotQA, while slightly underperforming on NQ.
However, we would like to highlight three key factors that can influence the performance:
(1) Training Data: We use only a subset of the HotpotQA training data, resulting in a zero-shot evaluation on the other datasets (NQ, TQA) for evaluating transfer capabilities.
(2) Model Weights: Unlike~\citet{cao2024retaining}, we do not fine-tune separate model weights for each dataset, demonstrating the versatility of our approach.
(3) Reranker Dependency: While~\citet{cao2024retaining} relies on a strong reranker (e.g., Cond.PPL~\citep{jiang2023longllmlingua}) to refine the input context, our approach does not rely on external rerankers, instead independently determining the priority of information within the context.

Notably, in HotpotQA, our framework surpasses the oracle setup reported in~\citet{cao2024retaining} where all gold documents are provided. 
Based on this, we hypothesize that providing complete gold evidence does not always create an optimal context for the reader model. 
Despite having the necessary information in the oracle, the F1 score remains significantly lower (57.7), indicating that the quality of summaries is critical. 
Our summaries, which prioritize essential information needed to answer the question, ensure that the context given to the reader model is both relevant and concise, leading to superior performance.
\begin{table}[h]
\centering
\resizebox{0.48\textwidth}{!}{
\begin{tabular}{lcccccc}
\toprule
\textbf{Methods} & \multicolumn{2}{c}{\textbf{NQ}} & \multicolumn{2}{c}{\textbf{TQA}} & \multicolumn{2}{c}{\textbf{HotpotQA}} \\
\cmidrule(lr){2-3} \cmidrule(lr){4-5} \cmidrule(lr){6-7}
 & \textbf{Comp.} & \textbf{Acc} & \textbf{Comp.} & \textbf{EM} & \textbf{Comp.} & \textbf{F1} \\
\midrule
Oracle & 59.2x & 73.5 & - & - & 42.2x & 57.7 \\
ICAE [2] & \textbf{21.5x} & 53.3 & 10.2x & 48.9 & 9.5x & 34.5 \\
QGC [1] & 15.2x & \textbf{60.9} & 7.9x & 57.5 & 8.8x & 51.6 \\
($\epsilon = 0.42$) & 20.6x & 57.6 & \textbf{10.9x} & 57.1 & 12.1x & 51.2 \\
CompAct (Ours) & 14.6x & 57.0 & \textbf{10.9x} & \textbf{64.4} & \textbf{12.2x} & \textbf{59.0} \\
\bottomrule
\end{tabular}
}
\caption{Comparison with ICAE~\citep{ge2024incontext} and QGC~\citep{cao2024retaining}. Following~\citet{cao2024retaining}, we use LLaMA2-7B as readers and report the Compression rate (Comp.) and F1 score.}
\label{tab:methods_comparison}
\end{table}

\section{Length of Compressed Text}
\paragraph{Token Length of Compressed Text per Iteration}
In Table~\ref{table:avg_length}, we provide detailed length information of compressed texts per iteration.
As the token length slightly increases with each iterations, We observe that \textsc{CompAct} maintains a high compression rate on average, which compresses 30 documents into under 200 tokens.
\begin{table}[h]
\centering
\resizebox{0.48\textwidth}{!}{
\begin{tabular}{l cccccc}
\toprule
\multicolumn{1}{c}{\multirow{2}{*}{\textbf{Datasets}}} & \multicolumn{6}{c}{\textbf{N-th Iterations}}                            \\ \cmidrule{2-7} 
\multicolumn{1}{c}{}                            & \multicolumn{1}{c}{\textbf{1}} & \multicolumn{1}{c}{\textbf{2}}   & \textbf{3}   & \multicolumn{1}{c}{\textbf{4}} & \multicolumn{1}{c}{\textbf{5}}   & \textbf{6} \\ \midrule
HotpotQA                                        & \multicolumn{1}{c}{78.1}      & \multicolumn{1}{c}{114.1} & 128.5 & \multicolumn{1}{c}{126.5}      & 135.9 & 147.5   \\ 
MuSiQue                 & \multicolumn{1}{c}{77.5}      & \multicolumn{1}{c}{110.6} & 135.2 & \multicolumn{1}{c}{91.6}      & \multicolumn{1}{c}{145.6}   &  124.0 \\ \bottomrule
\end{tabular}}{}
\caption{
Average token length of compressed texts per iteration.
5 documents are compressed for each iteration, as default setup of our framework.
}
\label{table:avg_length}
\end{table}

\paragraph{Token Usage}
In Table~\ref{table:token_usage}, we compare token usage against the baseline raw context setup with separate steps: Compress and Read. While \textsc{CompAct} generates more output tokens overall, it maintains a similar level of computational cost (see Table~\ref{table:computational_efficiency}). This is due to context segmentation, which mitigates the quadratic increase in computational cost associated with sequence length, resulting in significant computational efficiency benefits.
\begin{table}[h]
\centering
\resizebox{0.48\textwidth}{!}{
\begin{tabular}{lcccccc}
\toprule
\textbf{Dataset} & \textbf{Raw} & \multicolumn{3}{c}{\textbf{CompAct}} \\
\cmidrule(lr){2-2}
\cmidrule(lr){3-5}
 & \textbf{Total} & \textbf{Compress} & \textbf{Read} & \textbf{Total} \\
\midrule
HotpotQA  & 5218 / 7 & 4068 / 649 & 347 / 5 & 4415 / 654 \\
MusiQue   & 5125 / 8 & 5623 / 963 & 384 / 7 & 6007 / 970 \\
2WikiMQA  & 5453 / 8 & 4927 / 718 & 348 / 7 & 5275 / 724 \\
NQ        & 5023 / 8 & 2993 / 503 & 279 / 8 & 3273 / 511 \\
TQA       & 5091 / 5 & 2707 / 443 & 342 / 5 & 3049 / 447 \\
\bottomrule
\end{tabular}
}
\caption{Average token usage (input/output) per instance.}
\label{table:token_usage}
\end{table}

\section{Implementation Details}
\subsection{Training \& Inference}
\label{app:details}
We use 4 Nvidia A100 with 80GB memory to train our \textsc{CompAct} framework.
Our code is written in PyTorch~\citep{paszke2019pytorch} and HuggingFace~\citep{wolf2019huggingface}.
We use supervised fine-tuning through published alignment-handbook~\citep{alignment_handbook2023}. We train the model with Adam optimizer~\citep{KingBa15}, using a learning rate of 2e-6, a batch size of 64,  and 0.1 warm up ratio for 7 epochs.
For inference, all experiments are conducted using a greedy decoding strategy with a temperature of 0 and top\_p set to 1.0.

\subsection{Baselines}
\label{app:baselines}
\paragraph{Long-context LLMs.}
InternLM2-chat-7B~\cite{cai2024internlm2} has shown near-perfect performance on the Needle-in-the-Haystack task, which tests how well a model utilizes information within a long context.
Mistral-7B-Instruct-v0.2~\cite{jiang2023mistral} has recently shown strong performance across various benchmarks and supports a 32k context window.
FILM-7B~\cite{an2024make}, trained with a synthetic long-context question-answering dataset, has shown strong performance on tasks that require information awareness in the long context.
Phi-3-medium-128k-instruct and Phi-3.5-mini-instruct~\cite{phi}, leveraging their custom datasets, achieve state-of-the-art performance with a focus on high-quality reasoning.
Yi-9B-200k~\cite{yi} is an extended context version of the Yi series models.
Llama-3.1-8B-Instruct~\cite{llama3}, one of the latest models in the Llama family, supports a 128k context window, enabling enhanced performance on long-context tasks. We also experiment with GPT-3.5-turbo, a popular proprietary LLM that supports a 16k context window.

\paragraph{Compressors.}
(5) AutoCompressors~\cite{chevalier2023adapting} process segments of long context into soft prompts, which are prepended to the next segment as summary vectors. 
We use 50 summary tokens for every 2,048 tokens, following the setup from the original paper.
(6) LongLLMLingua~\cite{jiang2023longllmlingua} takes a perplexity-based approach to filter out tokens with less importance.
(7) RECOMP~\cite{xu2024recomp} suggests an extractive compressor that extracts relevant sentences using a dual encoder model, and an abstractive compressor that summarizes documents using an encoder-decoder model. 
We experiment with the extractive compressor setting, selecting 4 sentences from documents to ensure a fair comparison at similar text lengths.

\begin{figure}[t!]
\centering
\includegraphics[width=.78\columnwidth]{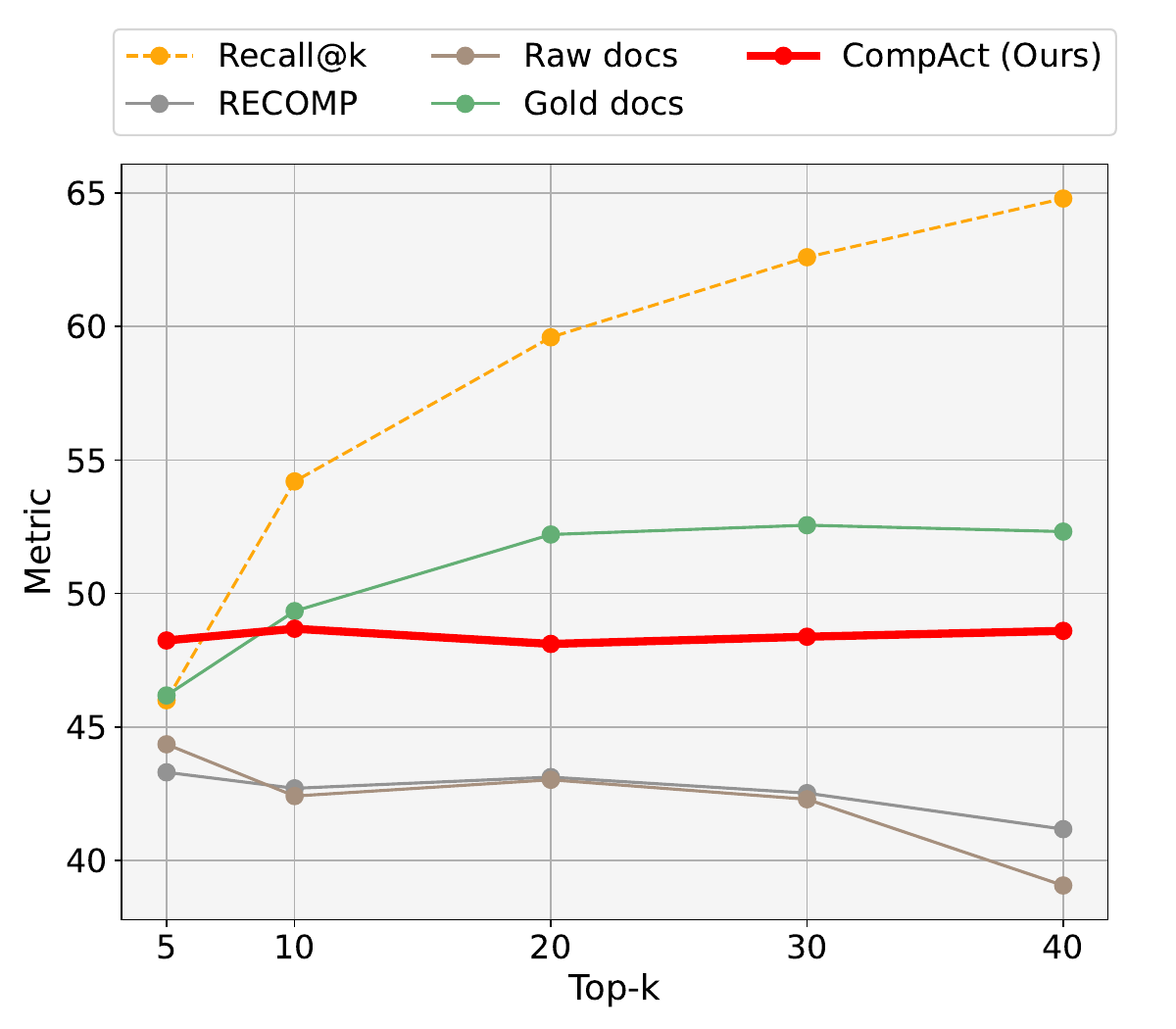} \\
\caption{
Performance of HotpotQA with different top-$k$ documents, using BM25 as the retriever.
}
\label{fig:bm25}
\vspace{-0.3cm}
\end{figure}
\begin{figure*}[]
\centering
\includegraphics[width=.98\columnwidth]{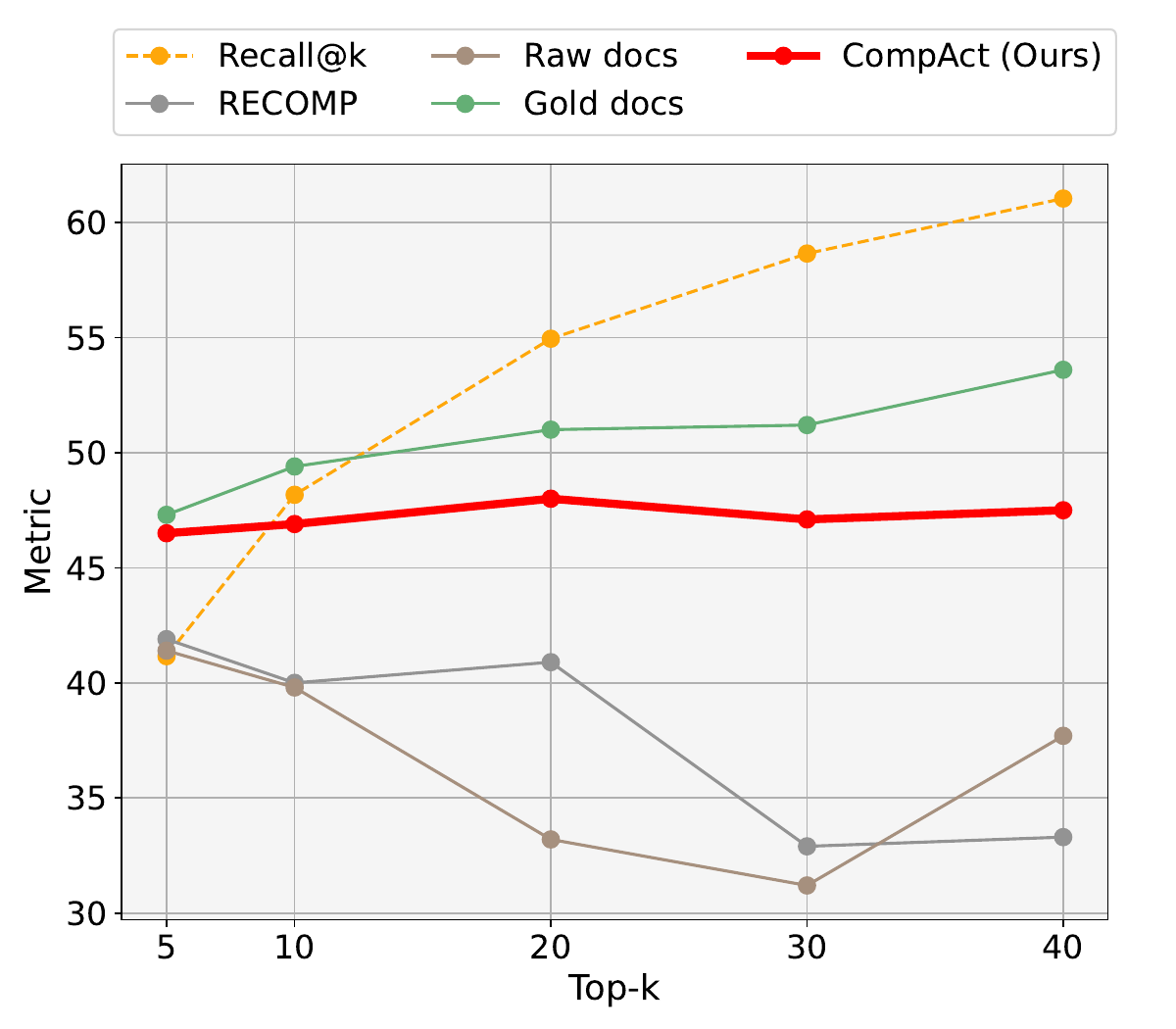}
\includegraphics[width=.98\columnwidth]{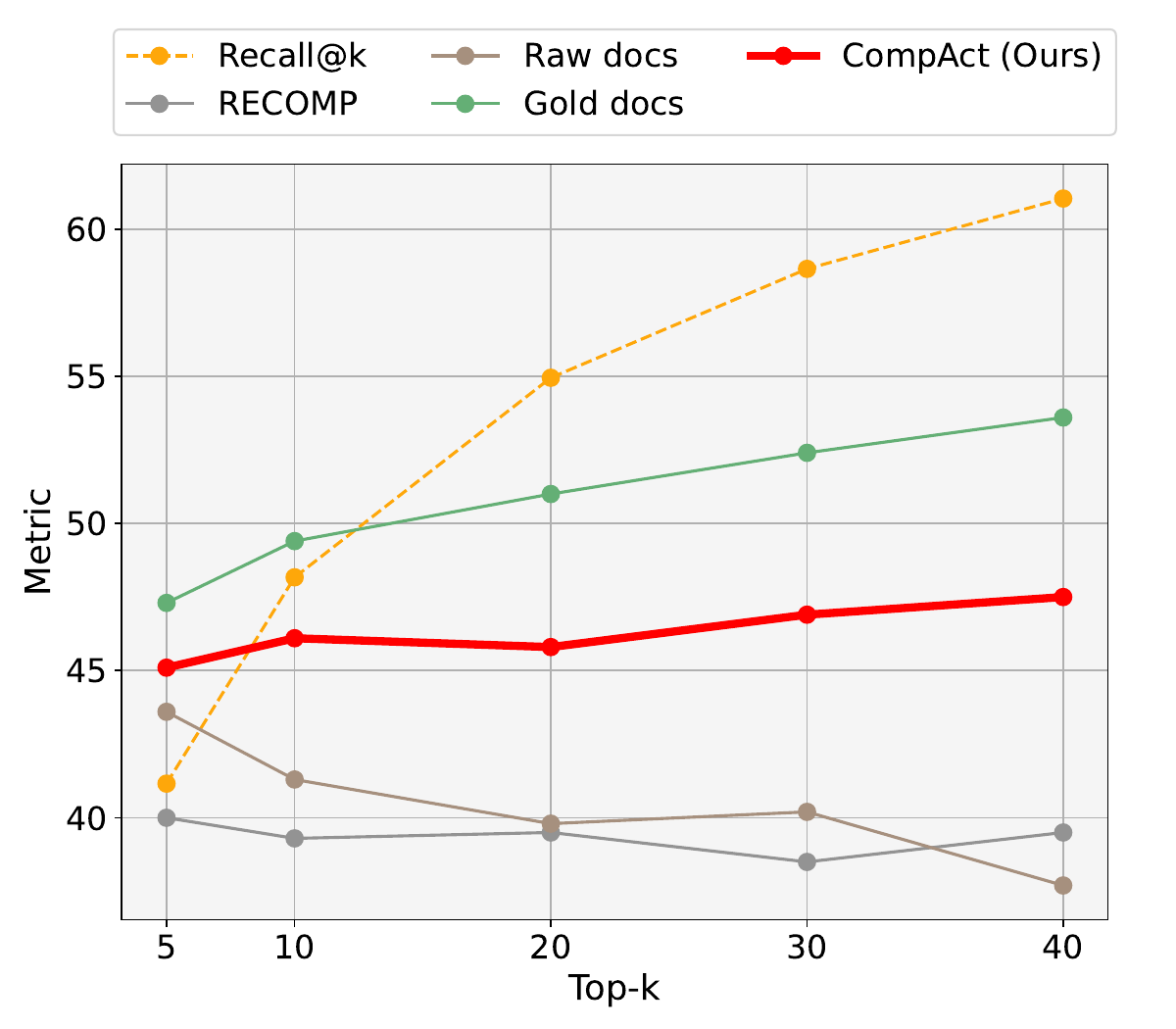}
\caption{
Performance of HotpotQA with different readers: LLaMA2-13B (left) and LLaMA3-8B (right).
}
\label{fig:reader_llama}
\end{figure*}
\begin{table*}[t]
\centering
{\resizebox{0.98\textwidth}{!}{
\begin{tabular}{l c c c c c}
\toprule
\textbf{Dataset}          & \textbf{Train}   & \textbf{Dev}    & \textbf{Test}   & \begin{tabular}[c]{@{}c@{}}\textbf{Avg. \# of} \\ \textbf{Supporting Documents}\end{tabular} & \begin{tabular}[c]{@{}c@{}}\textbf{\# of} \\ \textbf{Pre-defined Context}\end{tabular} \\ \midrule
NaturalQuestions~\citep{kwiatkowski2019natural} & 79,168  & 8,757  & 3,610  & - & - \\ 
TriviaQA~\citep{joshi-etal-2017-triviaqa}         & 78,785  & 8,837  & 11,313 & - & - \\ 
HotpotQA~\citep{yang2018hotpotqa}         & 90,447  & 7,405  & -      & 2 & 10 \\ 
MuSiQue~\citep{trivedi2022musique}          & 39,876  & 4,834  & 4,918  & 1.89 (Dev) & 20 \\ 
2WikiMultiHopQA~\citep{ho-etal-2020-constructing}         & 167,454 & 12,576 & 12,576 & 2.44 (Dev) & 10 \\ \bottomrule
\end{tabular}}}{}
\caption{Statistics of multi-hop and single-hop question answering datasets.}
\label{table:QA_statistics}
\end{table*}

\begin{table*}[t]
\footnotesize
\centering
\resizebox{\textwidth}{!}{
\begin{tabular}{p{0.95\textwidth}}
\toprule
\textbf{First Iteration:} \\\\
1. Generate a summary of source documents to answer the question. Ensure the summary is under 200 words and does not include any pronouns. DO NOT make assumptions or attempt to answer the question; your job is to summarize only. \\
\\
2. Evaluate the summary based solely on the information of it, without any additional background context: if it lacks sufficient details to answer the question, print [INCOMPLETE]. If it provides all necessary details, print [COMPLETE]. You should provide the reason of the evaluation. \\
\\
Question: [QUESTION] \\
\\
Source documents: [SOURCE DOCUMENTS] \\
\\
Summary: \\
\\
\midrule
\textbf{Subsequent Iterations:} \\\\
1. Generate a summary of the source documents and the previous summary to answer the question based on the evaluation of the previous summary. The evaluation indicates the missing information needed to answer the question. Ensure the summary is under 200 words and does not include any pronouns. DO NOT make assumptions or attempt to answer the question; your job is to summarize only. \\
\\
2. Evaluate the summary based solely on the information of it, without any additional background context: if it lacks sufficient details to answer the question, print [INCOMPLETE]. If it provides all necessary details, print [COMPLETE]. You should provide the reason of the evaluation. \\
\\
Question: [QUESTION] \\
\\
Evaluation of previous summary: [EVALUATION OF PREVIOUS SUMMARY] \\
\\
Previous summary: [PREVIOUS SUMMARY] \\
\\
Source documents: [SOURCE DOCUMENTS] \\
\\
Summary: \\
\\
\bottomrule
\end{tabular}}
\vspace{0.2cm}
\caption{Prompts used in \textsc{CompAct}}
\label{table:prompt_compact}
\end{table*}
\begin{table*}[t]
\footnotesize
\centering
\resizebox{\textwidth}{!}{
\begin{tabular}{p{0.95\textwidth}}
\toprule
Source sentences: [SOURCE SENTENCES] \\\\
Reference sentences: [REFERENCE SENTENCES] \\\\
Question: [QUESTION] \\ \\
Follow instructions below. \\
1. Choose 0 to 3 sentences that directly address the critical points needed to answer the question. Additionally, include 0 to 3 sentences that provide useful context, even if they do not directly answer the question. Ensure that you avoid selecting multiple sentences with overlapping content. (prefix: Sentences) \\
\\
2. Generate a summary of reference sentences and chosen sentences (prefix: Summary). Ensure the summary is under 200 words and does not include any pronouns. DO NOT make assumptions or attempt to answer the question; your job is to summarize only. \\
\\
3. Evaluate the summary based solely on the information of it, without any additional background context: if it lacks sufficient details to answer the question, print [INCOMPLETE]. If it provides all necessary details, print [COMPLETE]. You should provide the reason of evaluation (prefix: Evaluation) 
\\\\Sentences: \\



\bottomrule
\end{tabular}}

\vspace{0.2cm}
\caption{Prompt for dataset construction}
\label{table:prompt_dataset_construction}
\end{table*}
\colorlet{hlgreen}{green!30}
\colorlet{hlred}{red!30}

\newcommand{\hlgreen}[1]{{\sethlcolor{hlgreen}\hl{#1}}}
\newcommand{\hlred}[1]{{\sethlcolor{hlred}\hl{#1}}}
\begin{table*}[t]
\footnotesize
\centering{
\begin{tabular}{p{0.95\textwidth}}
\toprule
\\\textbf{Question}: Who was the eldest brother of the Mexican drug trafficker born 12 March 1952? \\ \\
\midrule
\textbf{Model}: \textsc{CompAct} (Ours) \\ 
\multicolumn{1}{c}{\textbf{Iter 1}} \\
\\
\textbf{Document (Rank 3)}: \hlgreen{Benjamín Arellano Félix (born 12 March 1952) is a Mexican drug trafficker} and former leader of the Mexican criminal organization known as the Tijuana Cartel or "Arellano-Félix Organization".
Benjamín Arellano Félix, who worked closely with his brothers, was one of Mexico's most powerful drug lords and the supplier of one-third of the U.S.'s cocaine. Benjamín had six brothers: He also has four sisters. 
Two of them, Alicia and Enedina, are most active in the cartel's affairs. Benjamín was first arrested on 18 June 1982, in Downey, California, for receiving 100 kilos of cocaine smuggled through the San\\ \\
\textbf{Compressed Text}: Benjamín Arellano Félix, born on 12 March 1952, is a Mexican drug trafficker and former leader of the Tijuana Cartel. He had six brothers, including He, and four sisters. He also has four sisters. \\ \\
\textbf{Evaluation}: The summary provides information about Benjamín Arellano Félix and his siblings but does not specify who the eldest brother is. \textbf{[INCOMPLETE]} \\ \\
\multicolumn{1}{c}{\textbf{Iter 2}} \\
\\
\textbf{Document (Rank 7)}: \hlgreen{Francisco Rafael Arellano Félix (24 October 1949 - 18 October 2013) was a Mexican drug lord and former leader of the Tijuana Cartel, a drug trafficking organization.
He was the oldest of seven brothers} and headed the criminal organization early in the 1990s alongside them. Through his brother Benjamín Francisco Rafael joined the Tijuana Cartel in 1989 following the arrest of Miguel Ángel Félix Gallardo, one of the most prominent drug czars in Mexico during the 1980s. When the Arellano Félix took control of the organization in the early 1990s, tensions with the rival Sinaloa Cartel\\ 
\\
\textbf{Compressed Text}: Francisco Rafael Arellano Félix, born on 24 October 1949, was a Mexican drug lord and former leader of the Tijuana Cartel. He was the oldest of seven brothers and headed the criminal organization early in the 1990s alongside them. Benjamín Arellano Félix, born on 12 March 1952, is a Mexican drug trafficker and former leader of the Tijuana Cartel. He had six brothers, including He, and four sisters. He also has four sisters. \\ \\
\textbf{Evaluation}: The summary provides all necessary details to answer the question. It identifies Francisco Rafael Arellano Félix as the oldest brother of Benjamín Arellano Félix, who was born on 12 March 1952. \textbf{[COMPLETE]} \\ \\
\textbf{Answer}: \hlgreen{Francisco Rafael Arellano Félix} (\textbf{Correct}) \\ \\
\midrule
\textbf{Model}: \textsc{Recomp}~\citep{xu2024recomp} \\ \\
\textbf{Summary}: Miguel Rodríguez Orejuela He is the younger brother of Gilberto Rodríguez Orejuela.\\ Roberto de Jesús Escobar Gaviria Roberto de Jesús Escobar Gaviria Roberto de Jesús Escobar Gaviria (born January 13, 1947), nicknamed El Osito (Little Bear), was the brother of the drug kingpin, Pablo Escobar, and the former accountant of the Medellín Cartel, which was responsible for up to 80 percent of the cocaine smuggled into the United States. \\ \\
\textbf{Answer}: \hlred{Fabio Ochoa Vásquez} (\textbf{Wrong}) \\ 
\bottomrule
\end{tabular}
}
\vspace{0.2cm}
\caption{Example of \textsc{CompAct} and comparison with \textsc{Recomp}}
\label{table:samples}
\end{table*}

\end{document}